\documentclass{article}

\usepackage[numbers]{natbib}
\usepackage[paper=8.5in:11in]{typearea}
\usepackage[margin=1.0in]{geometry}
\usepackage{lmodern}

\usepackage[utf8]{inputenc} \usepackage[T1]{fontenc}    \usepackage{hyperref}       \usepackage{url}            \usepackage{booktabs}       \usepackage{amsfonts}       \usepackage{nicefrac}       \usepackage{xcolor}         \usepackage{epsfig}
\usepackage{algorithm}
\usepackage{algorithmic}
\usepackage{amsthm}
\usepackage{amsmath}
\usepackage{float}
\usepackage{graphicx}
\usepackage{wrapfig}
\usepackage{makecell}
\usepackage{caption}
\usepackage{subcaption}
\usepackage{bm}
\usepackage{multirow}
\usepackage{microtype}
\usepackage{mathtools}
\usepackage{listings}
\usepackage{xspace}
\usepackage{amssymb}
\usepackage{tablefootnote}

\newcommand{\by}{{\mkern-2mu\times\mkern-2mu}}

\title{\vspace{-2em}\hrule height 4pt\vskip 0.25in\vskip -\parskip \textbf{
  On the Importance of Noise Scheduling for \\Diffusion Models
  }\vskip 0.2in\vskip -\parskip \hrule height 1pt\vskip 0.09in}

\author{
Ting Chen\\
Google Research, Brain Team \\
iamtingchen@google.com
}
\date{}

\begin{document}

\maketitle

\begin{abstract}
We empirically study the effect of noise scheduling strategies for denoising diffusion generative models.
There are three findings: (1) the noise scheduling is crucial for the performance, and the optimal one depends on the task (e.g., image sizes), (2) when increasing the image size, the optimal noise scheduling shifts towards a noisier one (due to increased redundancy in pixels), and (3) simply scaling the input data~\cite{chen2022generalist} by a factor of $b$ while keeping the noise schedule function fixed (equivalent to shifting the logSNR by $\log b$) is a good strategy across image sizes. This simple recipe, when combined with recently proposed Recurrent Interface Network (RIN)~\cite{jabri2022scalable}, yields state-of-the-art pixel-based diffusion models for high-resolution images on ImageNet, enabling \textit{single-stage, end-to-end} generation of diverse and high-fidelity images at 1024$\by$1024 resolution (without upsampling/cascades).
\end{abstract}

\begin{figure}[h!]
\begin{center}  
\includegraphics[width=0.85\linewidth]{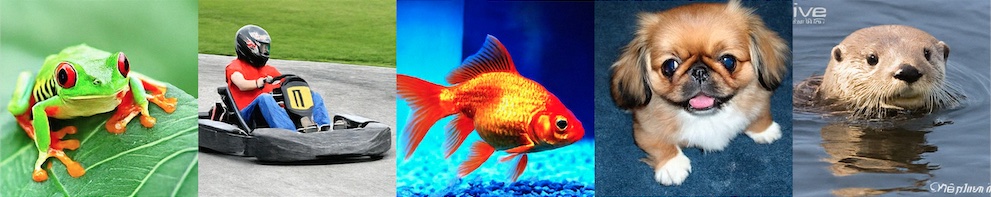}
\includegraphics[width=0.85\linewidth]{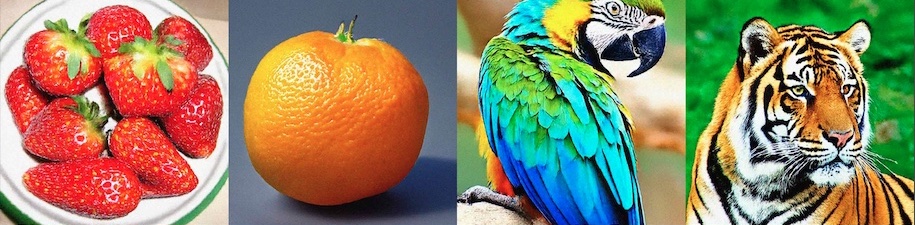}
\includegraphics[width=0.85\linewidth]{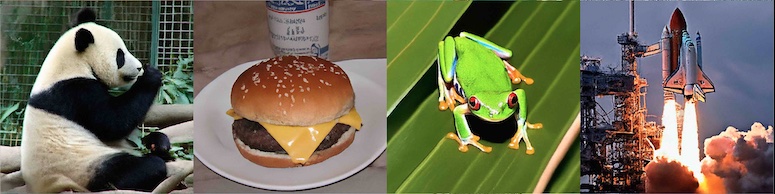}
\end{center}
\caption{\label{fig:samples_front} Random samples generated by our single-stage end-to-end model (trained on class-conditional ImageNet images) at high resolutions: $512\by512$ (the first row), $768\by768$ (the second row), $1024\by1024$ (the final row). More samples in Figure~\ref{fig:samples_512},~\ref{fig:samples_768} and~\ref{fig:samples_1024}.}
\end{figure} \section{Why is noise scheduling important for diffusion models?}

Diffusion models~\cite{sohl2015deep,ho2020denoising,song2020denoising,song2021scorebased,kingma2021variational,chen2022analog} define a noising process of data by $\bm x_t = \sqrt{\gamma(t)} \bm x_0 + \sqrt{1-\gamma(t)} \bm \epsilon$ where $\bm x_0$ is an input example (e.g., an image), $\bm \epsilon$ is a sample from a isotropic Gaussian distribution, and $t$ is a \textit{continuous} number between 0 and 1. The training of diffusion models is simple: we first sample $t\in \mathcal{U}(0, 1)$ to diffuse the input example $\bm x_0$ to $\bm x_t$, and then train a denoising network $f(\bm x_t)$ to predict either noise $\bm \epsilon$ or clean data $\bm x_0$. 
As $t$ is uniformly distributed, the noise schedule $\gamma(t)$ determines the distribution of noise levels that the neural network is trained on.

\begin{figure*}[h!]
\centering
\subfloat[$64\times 64$]{
\includegraphics[width=0.18\linewidth]{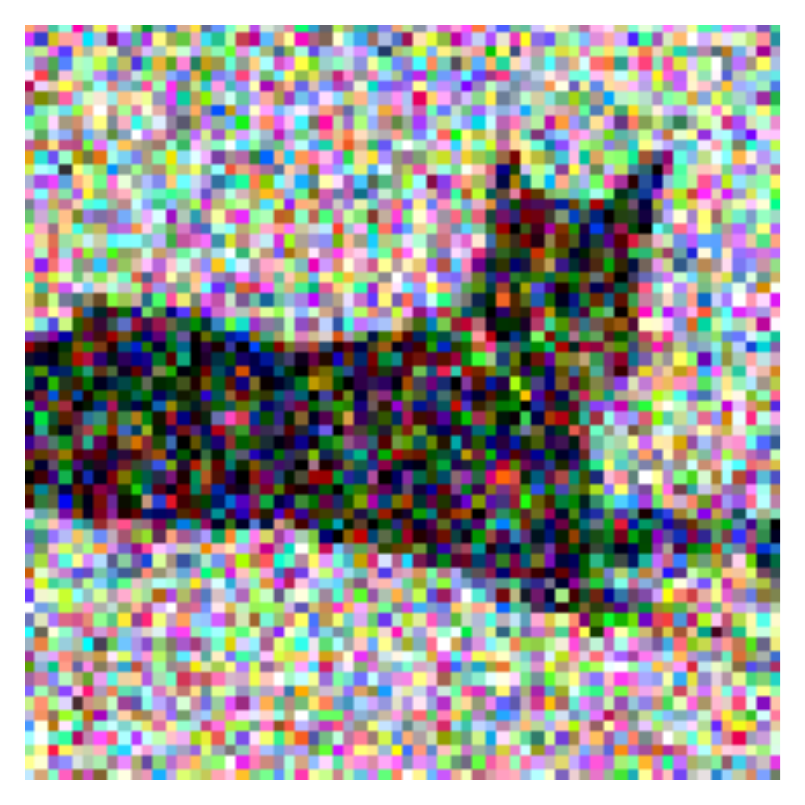} 
}
\subfloat[$128\times 128$]{
\includegraphics[width=0.18\linewidth]{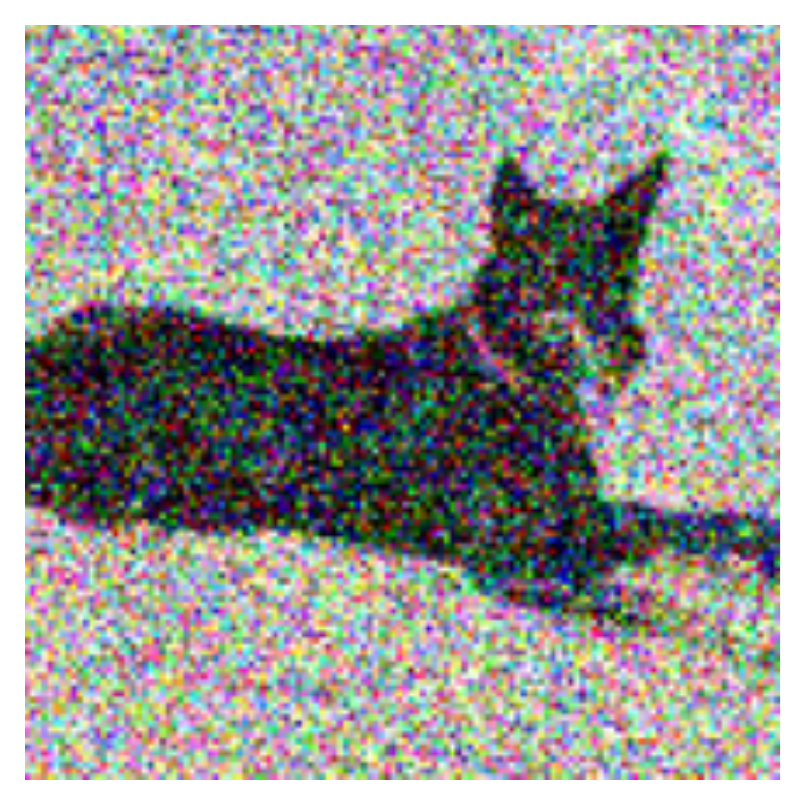}
}
\subfloat[$256\times 256$]{
\includegraphics[width=0.18\linewidth]{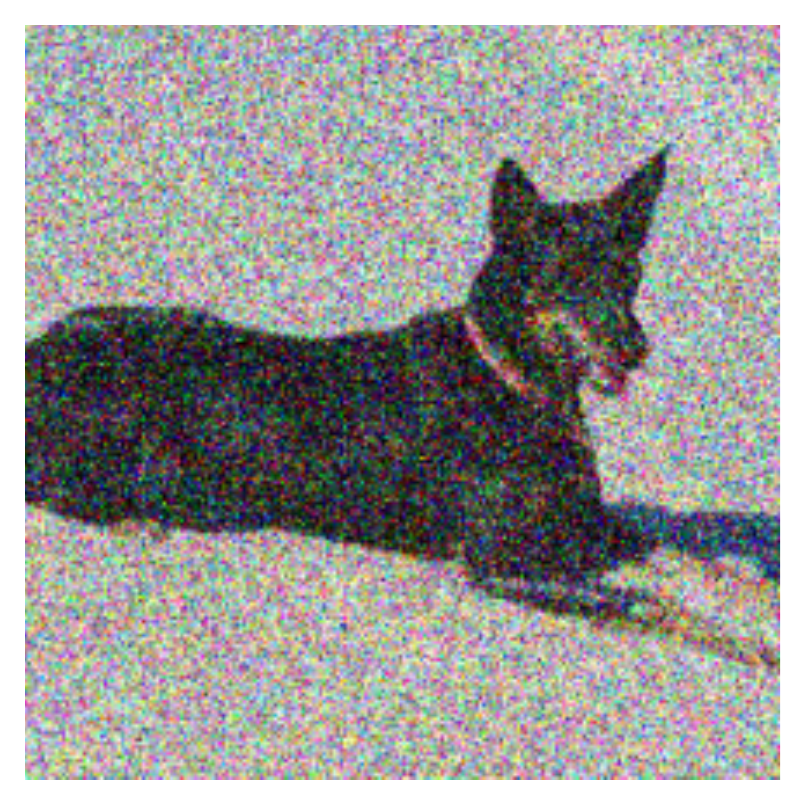}
}
\subfloat[$512\times 512$]{
\includegraphics[width=0.18\linewidth]{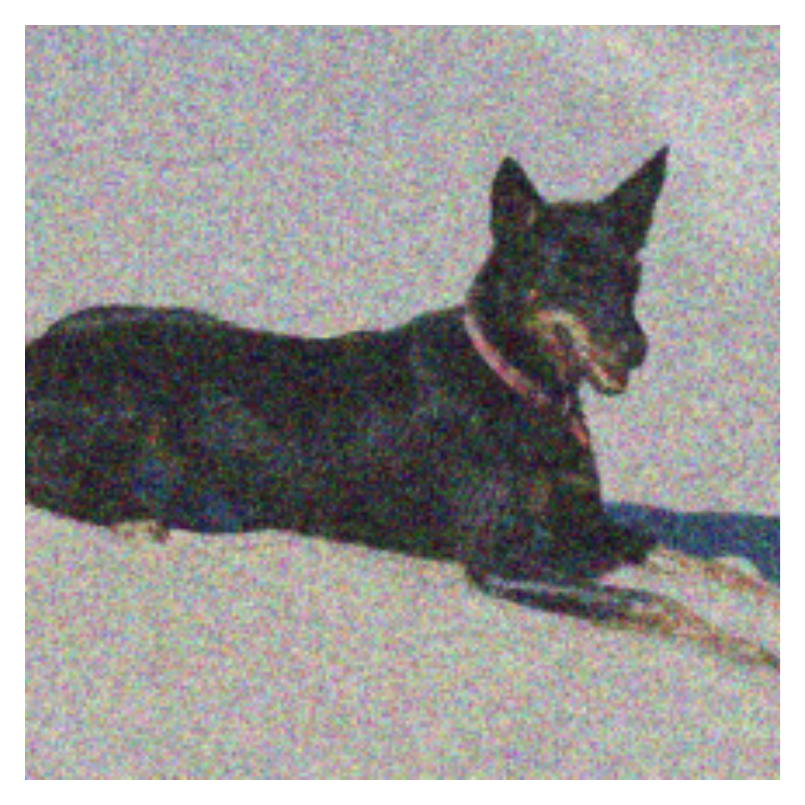}
}
\subfloat[$1024\times 1024$]{
\includegraphics[width=0.18\linewidth]{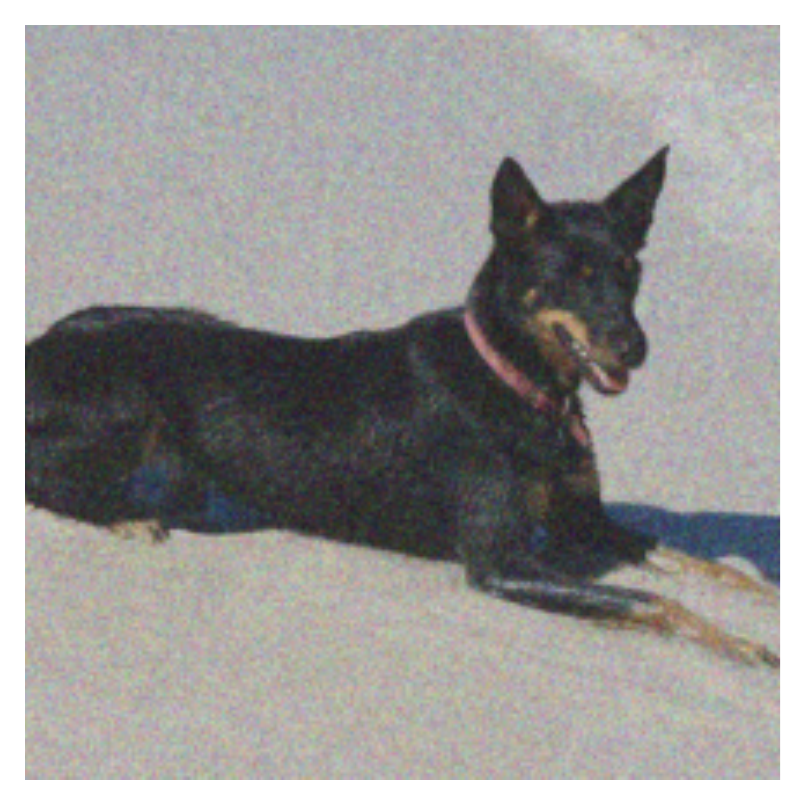}
}
\caption{\label{fig:noised_images}Noised images ($\bm x_t = \sqrt\gamma \bm x_0 + \sqrt{1-\gamma}\bm \epsilon$) with the same noise level ($\gamma=0.7$). We see that higher resolution natural images tend to exhibit higher degree of redundancy in (nearby) pixels, therefore less information is destroyed with the same level of independent noise. }
\end{figure*}

The importance of noise schedule can be demonstrated by the example in Figure~\ref{fig:noised_images}. As we increase the image size, the denoising task at the same noise level (i.e. the same $\gamma$) becomes simpler. This is due to the redundancy of information in data (e.g., correlation among nearby pixels) typically increases with the image size. Furthermore, the noises are independently added to each pixels, making it easier to recover the original signal when image size increases.
Therefore, the optimal schedule at a smaller resolution may not be optimal at a higher resolution. And if we do not adjust the scheduling accordingly, it may lead to under training of certain noise levels. Similar observations are made in concurrent work~\cite{simplediffusion,gu2022f}.

 \section{Strategies to adjust noise scheduling}

Built on top of existing work related to noise scheduling~\cite{ho2020denoising,nichol2021improved,kingma2021variational,chen2022generalist,jabri2022scalable}, we systematically study two different noise scheduling strategies for diffusion models.

\subsection{Strategy 1: changing noise schedule functions}

The first strategy is to parameterized noise schedule with a one-dimensional function~\cite{nichol2021improved,jabri2022scalable}. Here we present ones based on part of cosine or sigmoid functions, with temperature scaling. Note that the original Cosine schedule is proposed in~\cite{nichol2021improved}, with a fixed part of cosine curve that cannot be adjusted, and the simgoid schedule is proposed in~\cite{jabri2022scalable}. Other than these two types of functions, we further propose a simple linear noise schedule function, which is just $\gamma(t) = 1-t$ (note that this is not the linear schedule proposed in~\cite{ho2020denoising}).
Algorithm~\ref{alg:gamma} presents the code for these instantiations of the continuous time noise schedule function $\gamma(t)$.

\begin{figure}[ht]
  \centering
\begin{minipage}[t]{0.8\textwidth}
\begin{algorithm}[H]
\caption{Continuous time noise scheduling function $\gamma(t)$.
}
\label{alg:gamma}
\definecolor{codeblue}{rgb}{0.25,0.5,0.5}
\definecolor{codekw}{rgb}{0.85, 0.18, 0.50}
\lstset{
  backgroundcolor=\color{white},
  basicstyle=\fontsize{8.5pt}{8.5pt}\ttfamily\selectfont,
  columns=fullflexible,
  breaklines=true,
  captionpos=b,
  commentstyle=\fontsize{8.5pt}{8.5pt}\color{codeblue},
  keywordstyle=\fontsize{8.5pt}{8.5pt}\color{codekw},
  escapechar={|}, 
}
\begin{lstlisting}[language=python]
def simple_linear_schedule(t, clip_min=1e-9):
  # A gamma function that simply is 1-t.
  return np.clip(1 - t, clip_min, 1.)
  
def sigmoid_schedule(t, start=-3, end=3, tau=1.0, clip_min=1e-9):
  # A gamma function based on sigmoid function.
  v_start = sigmoid(start / tau)
  v_end = sigmoid(end / tau)
  output = sigmoid((t * (end - start) + start) / tau)
  output = (v_end - output) / (v_end - v_start)
  return np.clip(output, clip_min, 1.)
  
def cosine_schedule(t, start=0, end=1, tau=1, clip_min=1e-9):
  # A gamma function based on cosine function.
  v_start = math.cos(start * math.pi / 2) ** (2 * tau)
  v_end = math.cos(end * math.pi / 2) ** (2 * tau)
  output = math.cos((t * (end - start) + start) * math.pi / 2) ** (2 * tau)
  output = (v_end - output) / (v_end - v_start)
  return np.clip(output, clip_min, 1.)
\end{lstlisting}
\end{algorithm}
\end{minipage}
\end{figure}

\begin{figure*}[h!]
\centering
\subfloat[Cosine]{
\includegraphics[width=0.24\linewidth]{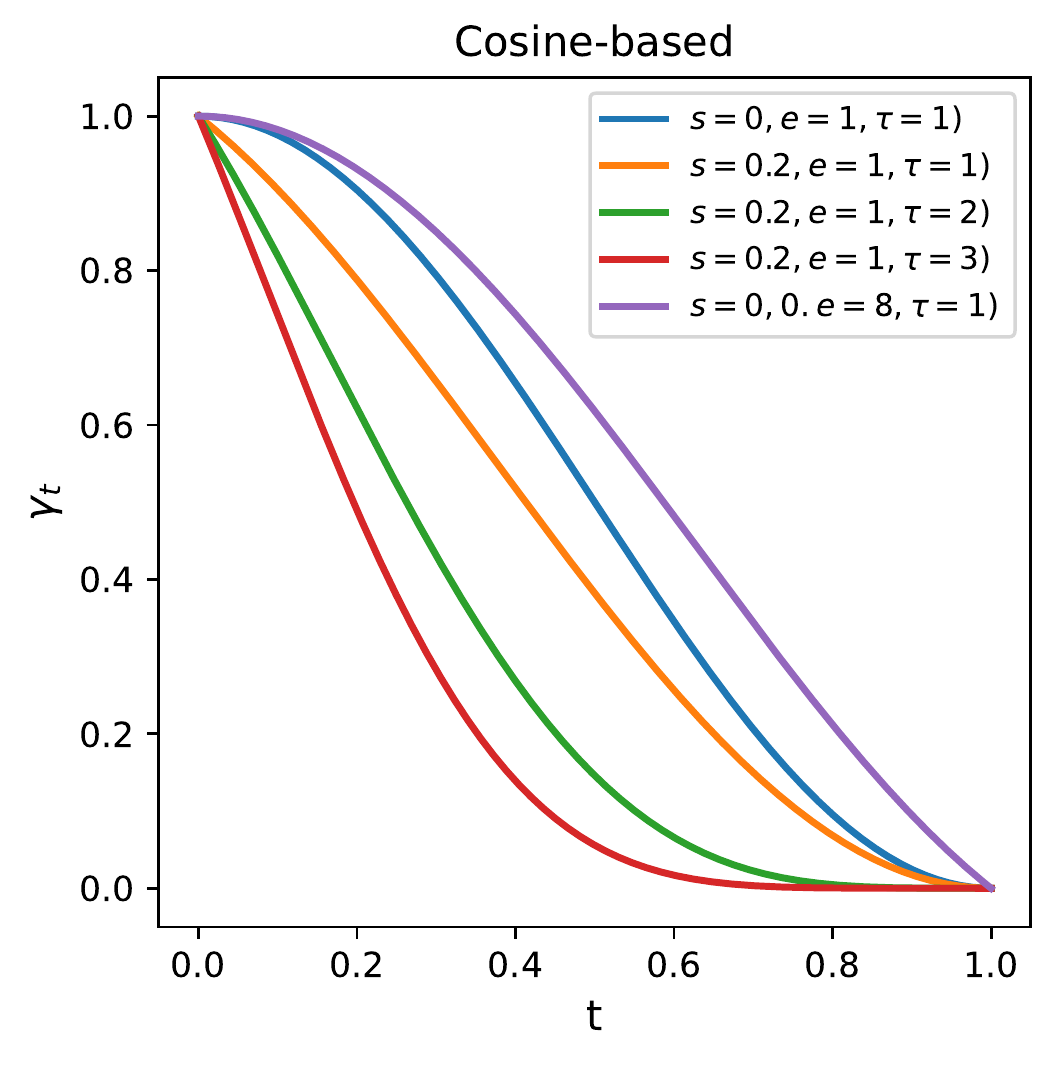} 
}
\subfloat[Cosine (logSNR)]{
\includegraphics[width=0.24\linewidth]{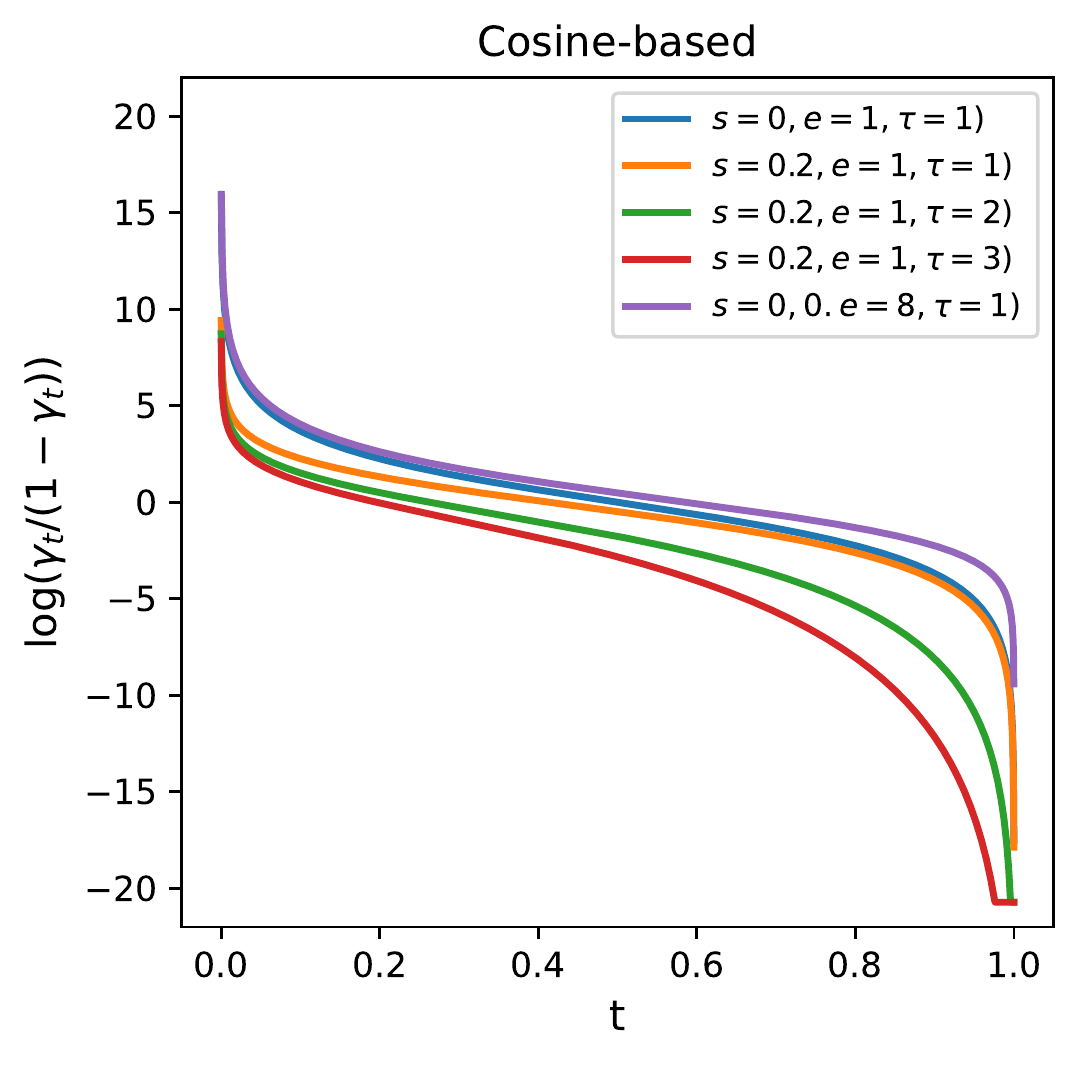}
}
\subfloat[Sigmoid]{
\includegraphics[width=0.24\linewidth]{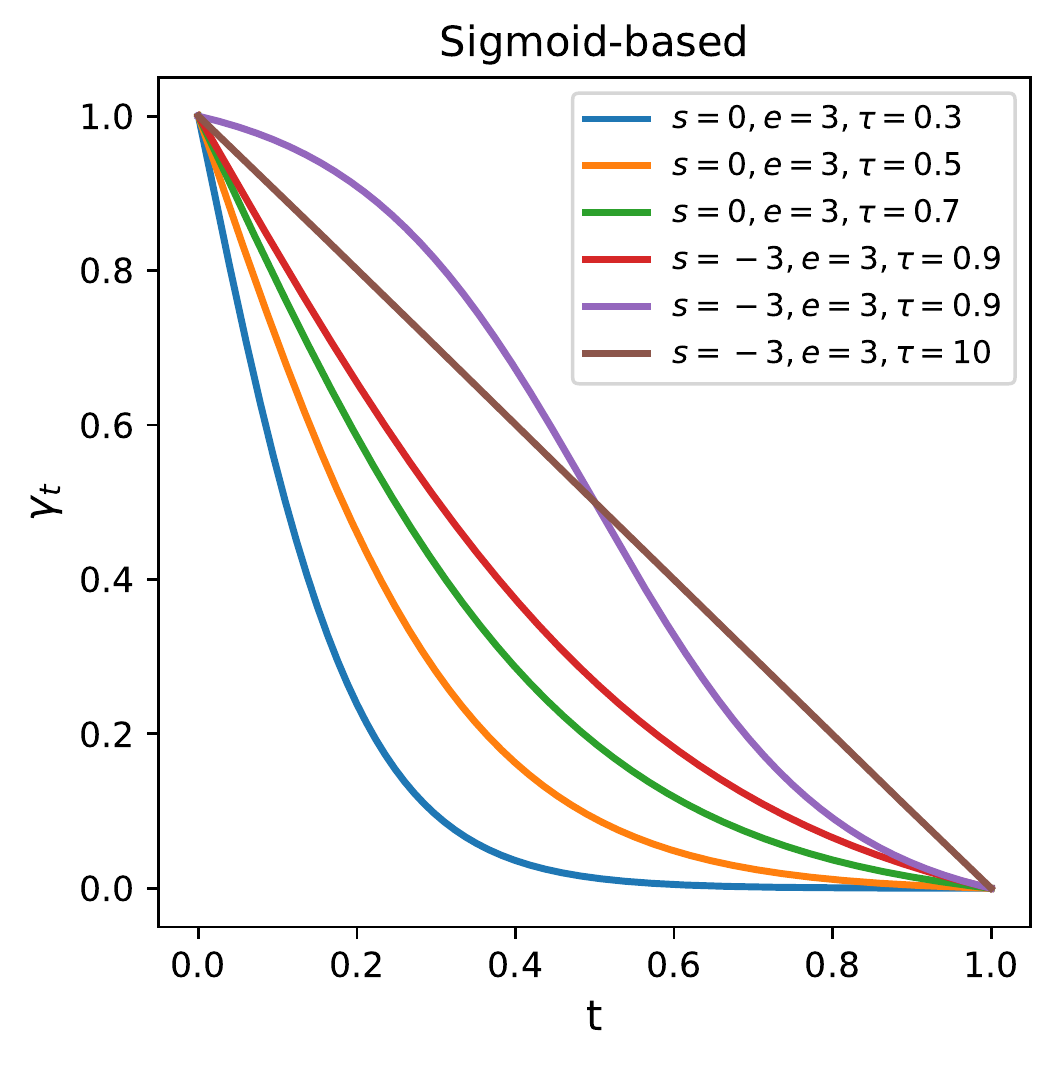}
}
\subfloat[Sigmoid (logSNR)]{
\includegraphics[width=0.24\linewidth]{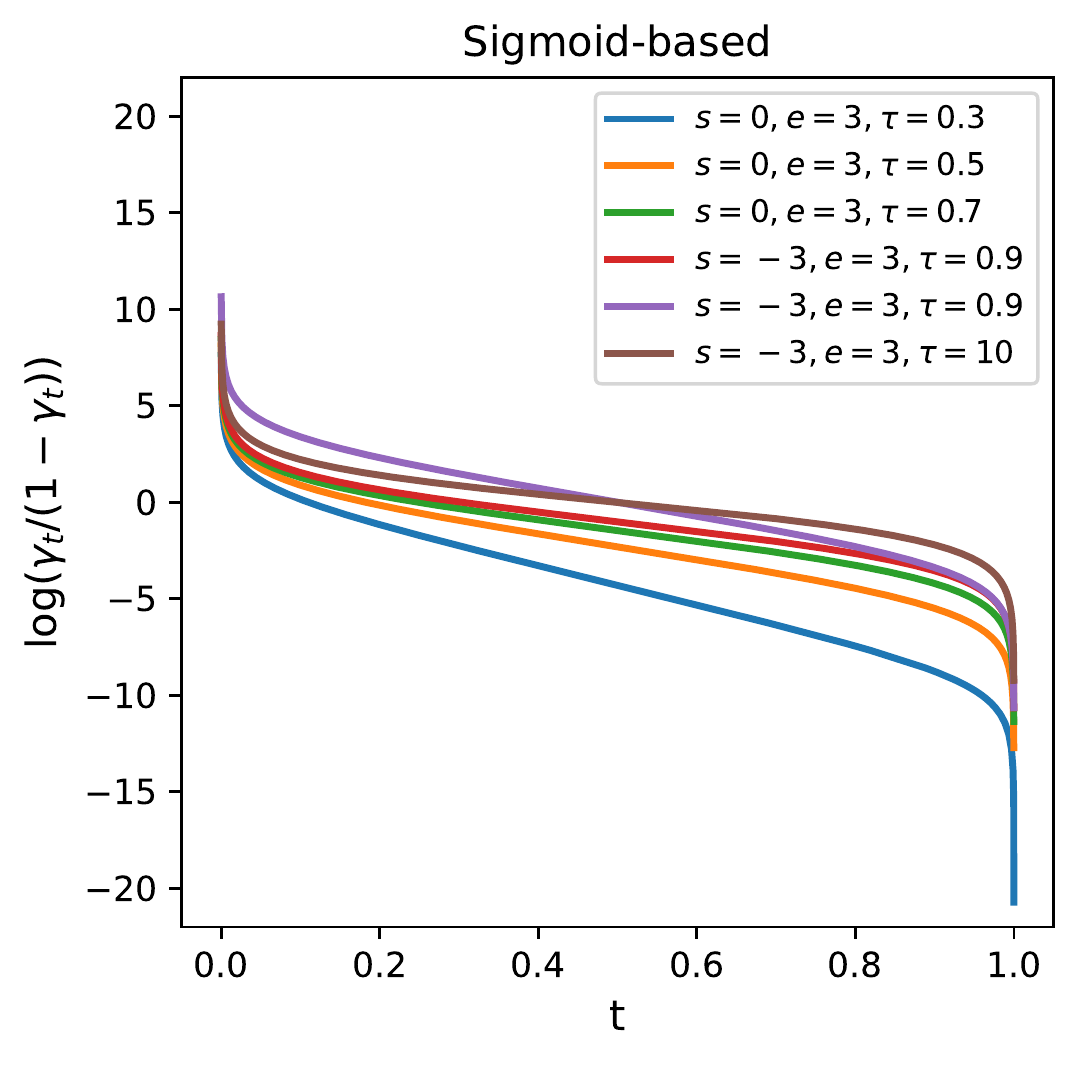}
}
\caption{\label{fig:noise_schedule}Instantiations of noise schedule function $\gamma(t)$ and the corresponding logSNR. Adjusting hyper-parameters of cosine and sigmoid functions leads to different noise schedules.}
\end{figure*}

Figure~\ref{fig:noise_schedule} visualizes the noise schedule functions under different choice of hyper-parameters, and their corresponding logSNR (signal-to-noise ratio). We can see that both cosine and sigmoid functions can parameterize a rich set of noise distributions. Please note that here we choose the hyper-parameters so that the noise distribution is skewed towards noisier levels, which we find to be more helpful.

\newpage
\subsection{Strategy 2: adjusting input scaling factor}

Another way to indirectly adjust noise scheduling, proposed in~\cite{chen2022generalist}, is to scale the input $\bm x_0$ by a constant factor $b$, which results in the following noising processing.
$$
\bm x_t = \sqrt{\gamma(t)} b \bm x_0 + \sqrt{1-\gamma(t)} \bm \epsilon
$$
\begin{figure*}[h!]
\begin{center}
\subfloat[$b=1$]{
\includegraphics[width=0.18\linewidth]{figures/noisy_image_512.pdf} 
}
\subfloat[$b=0.7$]{
\includegraphics[width=0.18\linewidth]{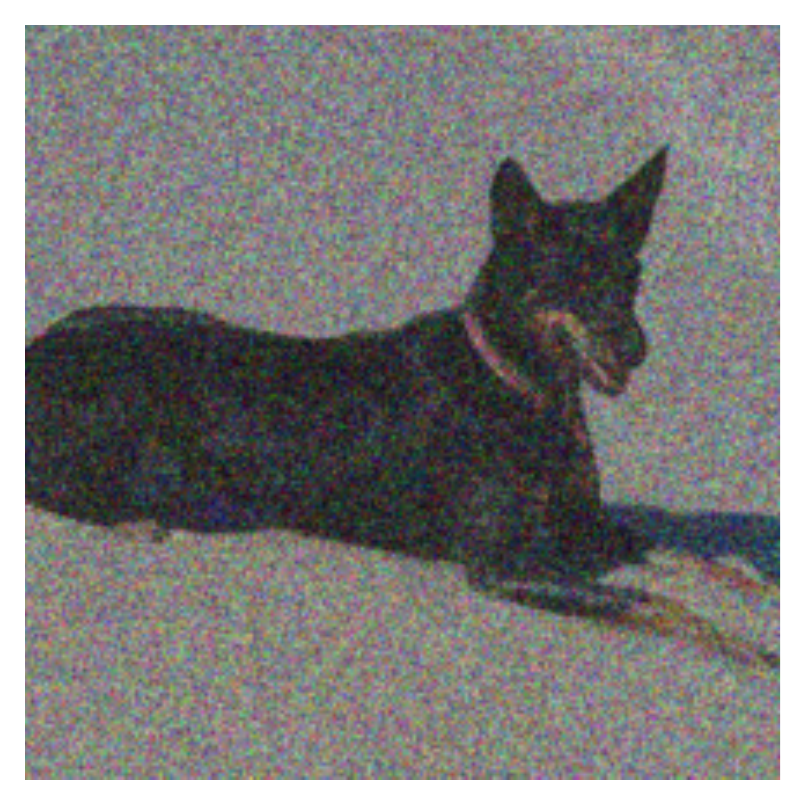}
}
\subfloat[$b=0.5$]{
\includegraphics[width=0.18\linewidth]{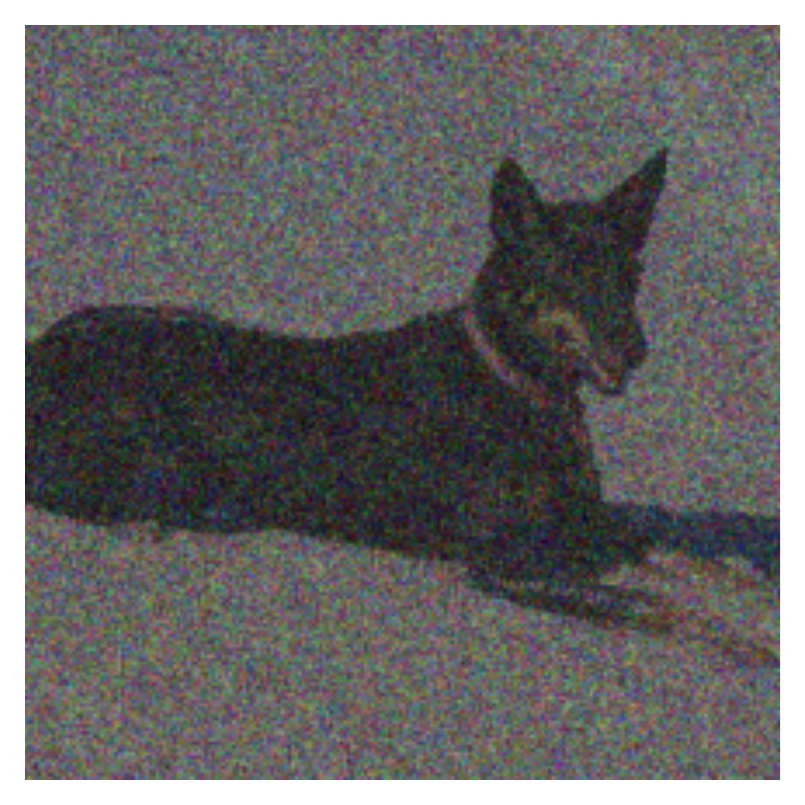}
}
\subfloat[$b=0.3$]{
\includegraphics[width=0.18\linewidth]{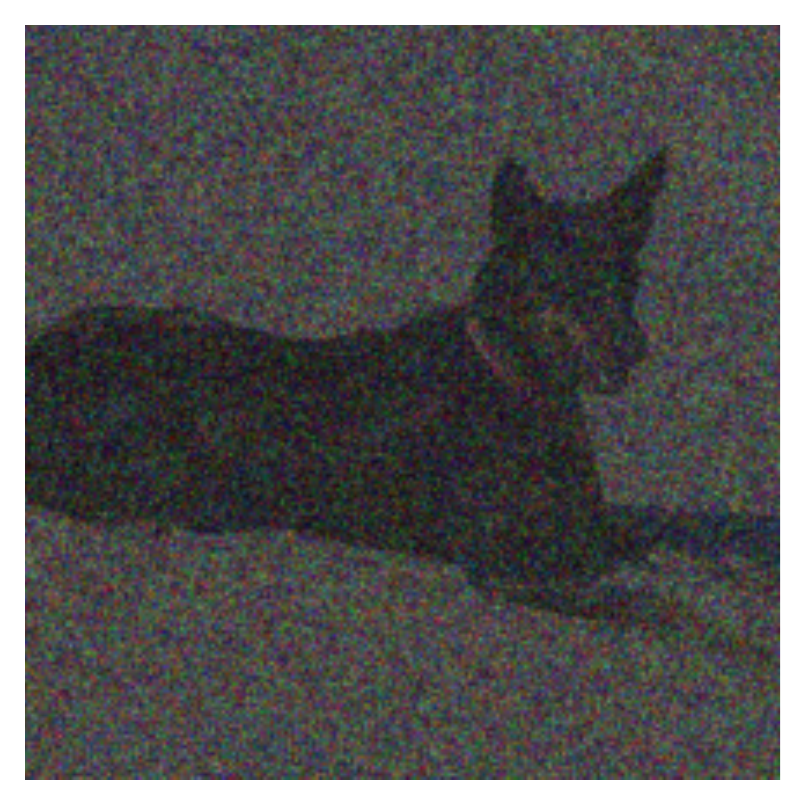}
}
\subfloat[$b=0.1$]{
\includegraphics[width=0.18\linewidth]{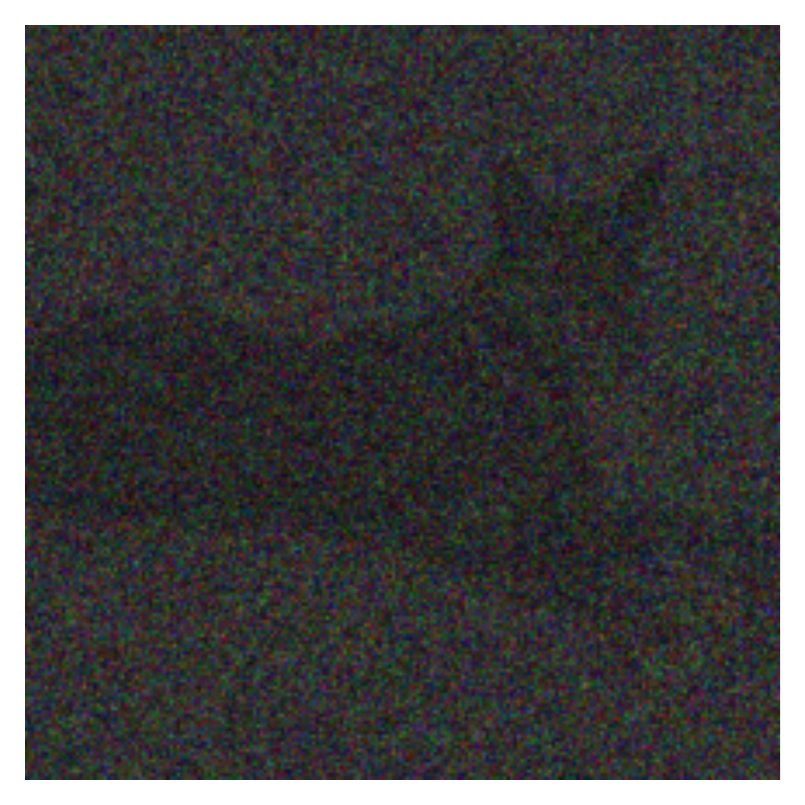}
}
\end{center}
\vspace{-1em}
\caption{\label{fig:noised_images_with_input_scaling}Noised images ($\bm x_t = \sqrt\gamma b\bm x_0 + \sqrt{1-\gamma}\epsilon$) with the same noise level ($\gamma=0.7$), but $\bm x_0$ is scaled by $b$. Using a smaller scaling factor, more information is destroyed with the same noise level. The noised image also becomes darker as the variance decreases.}
\end{figure*}
As we reduce the scaling factor $b$, it increases the noise levels, as demonstrated in Figure~\ref{fig:noised_images_with_input_scaling}.

When $b\neq1$, the variance of $\bm x_t$ can change even $\bm x_0$ has the same mean and variance as $\bm \epsilon$, which could lead to decreased performance~\cite{karras2022elucidating}. In this case, to ensure the variance keep fixed, one can scale $\bm x_t$ by a factor of $\frac{1}{(b^2-1)\gamma(t)+1}$. However, in practice, we find that it works well by simply normalize the $\bm x_t$ by its variance to make sure it has unit variance before feeding it to the denoising network $f(\cdot)$. This variance normalization operation can also be seen as the first layer of the denoising network.

\begin{figure*}[h!]
\centering
\includegraphics[width=0.42\linewidth]{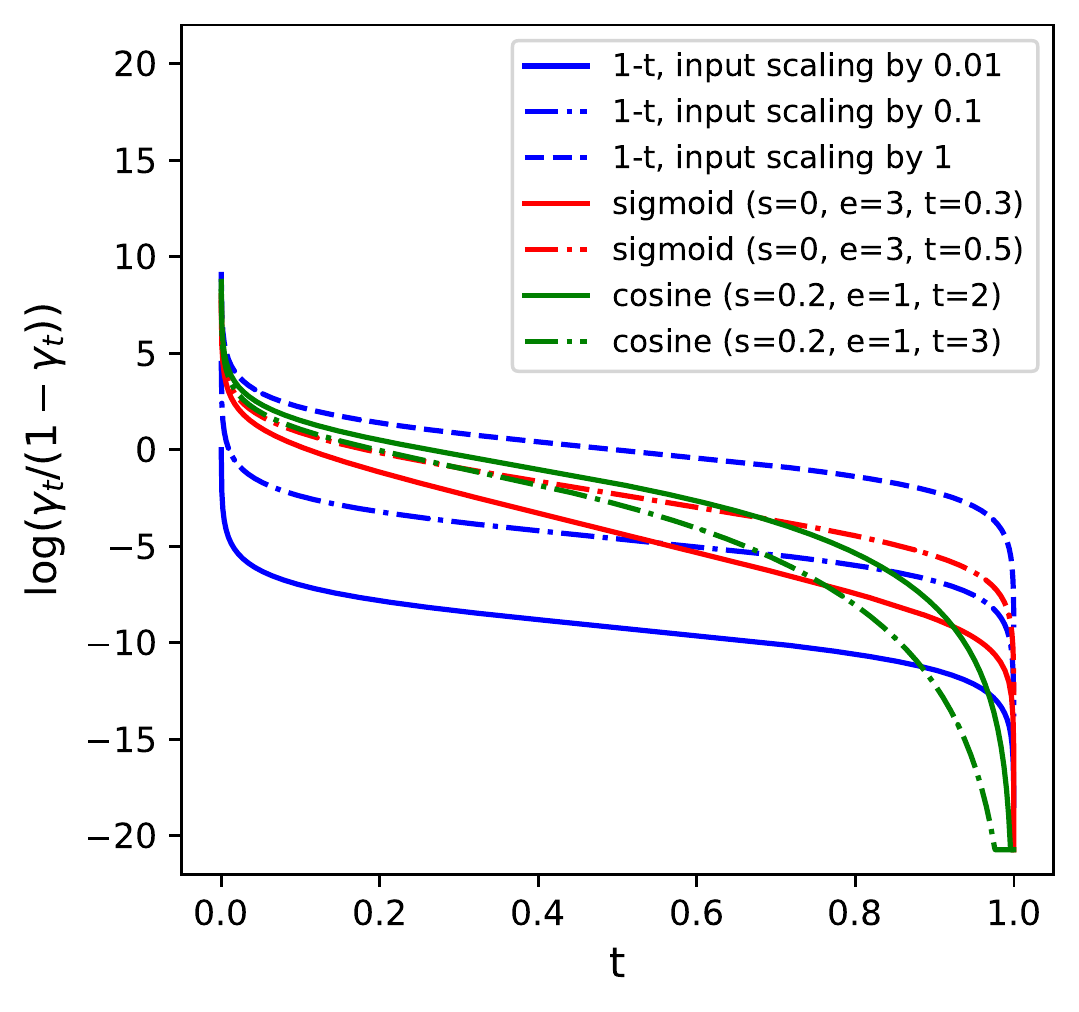}
\vspace{-0.5em}
\caption{\label{fig:noise_schedule2} Comparison of input scaling (on simple linear schedule) and other cosine-based or sigmoid based noise schedule functions. We can see the input scaling only shifts the logSNR along y-axis without changing its shape, while cosine and sigmoid functions put most emphasis on where $t$ is closer to 1, having much less influence when $t$ is smaller.}
\end{figure*}

While this input scaling strategy is similar to changing the noise scheduling function $\gamma(t)$ above, it achieves slightly different effect in the logSNR when compared to cosine and sigmoid schedules, particularly when $t$ is closer to 0, as shown in Figure~\ref{fig:noise_schedule2}. In fact, the input scaling shifts the logSNR along y-axis while keeping its shape unchanged, which is different from all the noise schedule functions considered above. Although, one may also equivalently parameterize $\gamma(t)$ function in other ways to avoid scaling the inputs, as nicely demonstrated by the concurrent work~\cite{simplediffusion}.

\subsection{Putting it together: a simple compound noise scheduling strategy}

Here we propose to combine these two strategies by having a single noise schedule function, such as $\gamma(t)=1-t$, and scale the input by a factor of $b$. The training and inference strategies are given in the following.

\paragraph{Training strategy}
Algorithm~\ref{alg:train} shows how to incorporate the combined noising scheduling strategy into the training of diffusion models, with main changes highlighted in blue.

\begin{algorithm}[H]
\caption{Training a diffusion model with the combined noise scheduling strategy.}
\label{alg:train}
\definecolor{codeblue}{rgb}{0.25,0.5,0.5}
\definecolor{codekw}{rgb}{0.85, 0.18, 0.50}
\lstset{
  backgroundcolor=\color{white},
  basicstyle=\fontsize{8.5pt}{8.5pt}\ttfamily\selectfont,
  columns=fullflexible,
  breaklines=true,
  captionpos=b,
  commentstyle=\fontsize{8.5pt}{8.5pt}\color{codeblue},
  keywordstyle=\fontsize{8.5pt}{8.5pt}\color{codekw},
  escapechar={|}, 
}
\vspace{-1mm}
\begin{lstlisting}[language=python]
def train_loss(x, gamma=lambda t: 1-t, scale=1, normalize=True):
  """Returns the diffusion loss on a training example x."""
  bsz, h, w, c = x.shape
  
  # Add noise to data.
  t = np.random.uniform(0, 1, size=[bsz, 1, 1, 1])
  eps = np.random.normal(0, 1, size=[bsz, h, w, c])
  x_t = np.sqrt(gamma(t)) * |\color{blue}scale *| x + sqrt(1-gamma(t)) * eps
  
  # Denoise and compute loss.
  |\color{blue}x\_t = x\_t / x\_t.std(axis=(1,2,3), keepdims=True) if normalize else x\_t|
  eps_pred = neural_net(x_t, t)
  loss = (eps_pred - eps)**2
  return loss.mean()
\end{lstlisting}
\vspace{-2mm}
\end{algorithm}

\paragraph{Inference/sampling strategy} If the variance normalization is used during the training, it should also be used during the sampling (i.e., the normalization can be seen as the first layer of the denoising network). Note that since we use a continuous time steps $t~\in[0, 1]$, so the inference schedule does not need to be the same as training schedule. During the inference we use a uniform discretization of the time between 0 and 1 into a given number of steps, and then we can chose a desired $\gamma(t)$ function to determine the level of noises at inference time. In practice, we find that standard cosine schedule works well for sampling.

\begin{algorithm}[H]
\small
\caption{\small Diffusion sampling algorithm.
}
\label{alg:sample}
\definecolor{codeblue}{rgb}{0.25,0.5,0.5}
\definecolor{codekw}{rgb}{0.85, 0.18, 0.50}
\lstset{
  backgroundcolor=\color{white},
  basicstyle=\fontsize{8.5pt}{8.5pt}\ttfamily\selectfont,
  columns=fullflexible,
  breaklines=true,
  captionpos=b,
  commentstyle=\fontsize{8.5pt}{8.5pt}\color{codeblue},
  keywordstyle=\fontsize{8.5pt}{8.5pt}\color{codekw},
  escapechar={|}, 
}
\vspace{-1mm}
\begin{lstlisting}[language=python]
def generate(steps, gamma=lambda t: 1-t, scale=1, normalize=True):
  x_t = normal(mean=0, std=1)
  for step in range(steps):
    # Get time for current and next states.
    t_now = 1 - step / steps
    t_next = max(1 - (step+1) / steps, 0)
        
    # Predict eps & jump to x at t_next.
    |\color{blue}x\_t = x\_t / x\_t.std(axis=(1,2,3), keepdims=True) if normalize else x\_t|
    eps_pred = neural_net(x_t, t_now)
    x_t = ddim_or_ddpm_step(x_t, eps_pred, t_now, t_next)
  return x_t
\end{lstlisting}
\vspace{-2mm}
\end{algorithm}
 \section{Experiments}

\subsection{Setup}
\label{sec:train_details}

We mainly conduct experiments on class-conditional ImageNet~\cite{russakovsky2015imagenet} image generation, and we follow common practice of evaluation, using FID~\cite{heusel2017gans} and Inception Score~\cite{salimans2016improved} as metrics computed on 50K samples, generated by 1000 steps of DDPM.

We follow~\cite{jabri2022scalable} for model specification but use smaller models as well as shorter overall training steps (except for >256 resolutions) to conserve compute. This results in worse performance in general but due to the improvement of noise scheduling, we can still achieve similar performance at lower resolutions (64$\by$64 and 128$\by$128), but significantly better results at higher resolutions (256$\by$256 or higher).

For hyper-parameters, we use LAMB~\cite{you2019large} optimizer with $\beta_1=0.9, \beta_2=0.999$ and weight decay of 0.01, self-conditioning rate of 0.9, and EMA decay of 0.9999. Table~\ref{tab:model_hypers} and ~\ref{tab:train_hypers} summarize major hyper-parameters.

\begin{table}[h]
\small
\centering
\caption{\label{tab:model_hypers} Model Hyper-parameters.}
\vspace{-0.5em}
\begin{tabular}{ccccccccc}
\toprule
Image Size & Patch Size & Tokens & Latents & Layers & Heads & Params & Input Scale & $\gamma(t)$ \\
\midrule
$64\by 64 \by 3$ & $8\by 8$ & 64$\by$512 & 128$\by$768 & 6,6,6,6 & 16 & 214M & 1.0 & 1-t\\
$128\by 128 \by 3$ & $8\by 8$ & 256$\by$512 & 128$\by$768 & 6,6,6,6 & 16 & 215M & 0.6 & 1-t\\
$256\by 256 \by 3$ & $8\by 8$ & 1024$\by$512 & 256$\by$768 & 6,6,6,6,6,6 & 16 & 319M & 0.5 & 1-t\\
$512\by 512 \by 3$ & $8\by 8$ & 4096$\by$512 & 256$\by$768 & 6,6,6,6,6,6 & 16 & 320M & 0.2 & cosine@0.2,1,1~\tablefootnote{Here $\gamma(t)=1-t$ should work as well but it is not compared in our limited experiments.}\\
$768\by 768 \by 3$ & $8\by 8$ & 9216$\by$512 & 256$\by$768 & 8,8,8,8,8,8 & 16 & 408M & 0.1 & 1-t\\
$1024\by1024\by3$ & $8\by 8$ & 16384$\by$512 & 256$\by$768 & 8,8,8,8,8,8 & 16 & 412M & 0.1 & 1-t\\
\bottomrule
\end{tabular}
\end{table}

\begin{table}[h!]
\centering
\small
\caption{\label{tab:train_hypers}Training Hyper-parameters.}
\vspace{-0.5em}
\begin{tabular}{cccccc}
\toprule
Image Size & Train Steps & Batch Size & LR & LR Decay & Label Dropout\\
\midrule
$64\by 64 \by 3$  & 150K & 1024 & 2e-3 & Cosine (first 70\%) & 0.0\\
$128\by 128 \by 3$  & 250K & 1024 & 2e-3 & Cosine (first 70\%) & 0.0\\
$256\by 256 \by 3$  & 250K & 1024 &2e-3 & Cosine (first 70\%) & 0.0\\
$512\by 512 \by 3$  & 1M & 1024 &1e-3 & Constant & 0.0\\
$768\by 768 \by 3$  & 1M & 1024 &1e-3 & Constant & 0.1\\
$1024\by 1024 \by 3$  & 910K & 1024 &1e-3 & Constant & 0.1\\
\bottomrule
\end{tabular}
\end{table}

\subsection{The effect of strategy 1 (noise schedule functions)}

We first keep the input scaling fixed to 1, and evaluate the effect of noise schedules based on cosine, sigmoid and linear functions. As shown in Table~\ref{tab:exp_noise_schedule}, \textbf{different image resolutions require different noise schedule functions to obtain the best performance, and it is difficult to find the optimal schedule due to several hyper-parameters involved}.

\begin{table}[h]
    \centering
    \small
    \caption{\label{tab:exp_noise_schedule}FIDs for different noise schedule functions (see Figure~\ref{fig:noise_schedule} for visualization) while keeping the input scaling fixed to 1. For FID, the lower the better. For different image resolutions, optimal schedule function is quite different, making it difficult to find/tune.}
\begin{tabular}{c c c c}
    \toprule
    Noise schedule function $\gamma(t)$ & $64\by64$ & $128\by128$ & $256\by256$ \\
    \midrule
    1-t & \textbf{2.04} & 4.51 & 7.21 \\ 
    \midrule
    cosine (s=0,e=1,$\tau=1$; i.e., cosine) & 2.71 & 7.28 & 21.6 \\
    cosine (s=0.2,e=1,$\tau=1$) & 2.15 & 4.9 & 12.3 \\
    cosine (s=0.2,e=1,$\tau=2$) & 2.84 & 5.64 & 5.61 \\
    cosine (s=0.2,e=1,$\tau=3$)  & 3.3 & 4.64 & 6.24 \\
    \midrule
    sigmoid (s=-3,e=3,$\tau=0.9$) & 2.09 & 5.83 & 7.19 \\
    sigmoid (s=-3,e=3,$\tau=1.1$) & \textbf{2.03} & 4.89 & 7.23 \\
    sigmoid (s=0,e=3,$\tau=0.3$) & 4.93 & 6.07 & 5.74 \\
    sigmoid (s=0,e=3,$\tau=0.5$) & 3.12 & 5.71 & \textbf{4.28} \\
    sigmoid (s=0,e=3,$\tau=0.7$) & 3.34 & \textbf{3.91} & 5.49 \\
    sigmoid (s=0,e=3,$\tau=0.9$) & 2.29 & 4.42 & 5.48 \\
    sigmoid (s=0,e=3,$\tau=1.1$) & 2.36 & 4.39 & 7.15 \\ 
    \bottomrule
    \end{tabular}
\end{table}

\subsection{The effect of strategy 2 (input scaling)}

\begin{table}[h!]
    \centering
    \small
    \caption{\label{tab:exp_scaling}FIDs for different input scaling factors while keeping the noise schedule function fixed to either cosine (s=0.2,e=1,$\tau=1$) or $1-t$. For FID, the lower the better.}
    \vspace{-0.5em}
\begin{tabular}{c c c c c c c}
    \toprule
    \multirow{2}{*}{Input scale factor} & \multicolumn{2}{c}{$64\by64$} & \multicolumn{2}{c}{$128\by128$} & \multicolumn{2}{c}{$256\by256$} \\
    \cmidrule(lr){2-3} \cmidrule(lr){4-5} \cmidrule(lr){6-7}
    & cosine@0.2,1,1 & $1-t$ & cosine@0.2,1,1 & $1-t$ & cosine@0.2,1,1 & $1-t$ \\
    \midrule
    0.3 & 5.1 & 6.77 & 5.63 & 5.25 & \textbf{3.7} & \textbf{3.58} \\
    0.4 & 4 & 3.79 & 4.65 & 6.89 & 4.01 & \textbf{3.52} \\
    0.5 & 3.76 & 3.79 & 4.14 & 3.9 & 5.12 & 5.07 \\
    0.6 & 3.42 & 2.8 & \textbf{3.97} & \textbf{3.5} & 5.54 & 5.54 \\
    0.7 & 2.4 & 2.49 & 4.78 & 5.34 & 7.93 & 5.72 \\
    0.8 & 2.36 & 2.43 & 6.28 & 5.35 & 4.52 & 7.52 \\
    0.9 & 2.31 & 2.23 & 4.89 & 3.86 & 5.51 & 6.69 \\
    1 & \textbf{2.15} & \textbf{2.04} & 4.9 & 4.51 & 12.3 & 7.21 \\
    \bottomrule
    \end{tabular}
\end{table}

Here we keep the noise schedule functions fixed, and adjust the input scaling factor. The results are shown in Table~\ref{tab:exp_scaling}. We find that \textbf{1) as image resolution increases, the optimal input scaling factor becomes smaller, 2) compared to the best result from Table~\ref{tab:exp_noise_schedule} where we only change the noise schedule function while keeping input scaling fixed, adjusting input scaling is better (drop FID from 4.28 to 3.52 for $256\by256$), and it is also easier to find as we can just tune a single scaling factor}. Finally, $1-t$ seems to be a slightly better noise schedule than cosine (s=0.2,e=1,$\tau=1$).

\subsection{The simple compound strategy, combined with RIN~\cite{jabri2022scalable}, enables state-of-the-art single-stage high-resolution image generation based on pixels}

Table~\ref{tab:image_sota} demonstrates that the simple compound noise scheduling strategy, combined with RIN~\cite{jabri2022scalable}, enables state-of-the-art generation of high resolution images based on pure pixels. We forgo latent diffusion models~\cite{rombach2022high} where ``pixels'' are replaced with learned latent codes, since our scheduling technique is only tested on pixel-based diffusion models, but note these are orthogonal techniques and can potentially be combined.
We note that state-of-the-art GANs~\cite{sauer2022stylegan} can achieve similar or better performance but with multi-stage generation, as well as classifier-guidance~\cite{dhariwal2021diffusion}, which we do not use for quantitative evaluation. 

\begin{table}[h!]
\centering
\small
\caption{\label{tab:image_sota}Comparison of state-of-the-art class-conditional pixel-based image generation models on ImageNet. For FID, the lower the better; for IS, the higher the better. Our results (based on RIN) reported use neither cascades/up-sampling nor guidance.}
\vspace{-0.5em}
\begin{tabular}{clcccc}
\toprule
Resolution & Method & FID & IS & Params (M) \\
\midrule

\multirow{5}{*}{$64\by 64$} &
ADM~\cite{dhariwal2021diffusion}  & -- &  2.07 & 297 \\
&CF-guidance~\cite{ho2021classifierfree}  & 1.55 & 66.0 & --\\
&CDM~\cite{ho2021cascaded}  & 1.48 & 66.0 &  -- \\
&RIN~\cite{jabri2022scalable} (patch size of 4, 300K updates)  & \textbf{1.23} & \textbf{66.5} & 281 \\
&RIN+our strategy (patch size of 8, 150K updates)  & 2.04 & 55.8 & \textbf{214} \\
\midrule

\multirow{6}{*}{$128\by 128$} &
ADM~\cite{dhariwal2021diffusion}  & 5.91 & -- & 386 \\
&ADM+guidance~\cite{dhariwal2021diffusion}  & 2.97 & -- &  > 386\\
&CF-guidance\cite{ho2021classifierfree}  & \textbf{2.43} & \textbf{156.0} &  --  \\
&CDM\cite{ho2021cascaded}  & 3.51 & 128.0 & 1058  \\
&RIN~\cite{jabri2022scalable} (patch size of 4, 700K updates) & 2.75 & 144.1 & 410 \\
&RIN+our strategy (patch size of 8, 250K updates) & 3.50 & 120.4 &  \textbf{215} \\
\midrule

\multirow{5}{*}{$256\by 256$} &
ADM~\cite{dhariwal2021diffusion}  & 10.94 & 100.9 &  553 \\
&ADM+guidance~\cite{dhariwal2021diffusion}  & 4.59 & -- & >553\\
&CDM~\cite{ho2021cascaded}  & 4.88 & 158.7 & 1953\\
&RIN~\cite{jabri2022scalable} (patch size of 8, 700K updates) & {4.51} & {161.0} & 410 \\
&RIN+our strategy (patch size of 8, 250K updates)  & \textbf{3.52} & \textbf{186.2} & \textbf{319} \\
\midrule

\multirow{1}{*}{$512\by 512$} &
ADM~\cite{dhariwal2021diffusion}  & 23.2 & 58.1 &  559 \\
&ADM+guidance~\cite{dhariwal2021diffusion}  & 7.72 & 172.7 & >559\\
&RIN+our strategy (patch size of 8, 1M updates)  & \textbf{3.95} & \textbf{216} & \textbf{320} \\
\midrule

\multirow{1}{*}{$768\by 768$}
&RIN+our strategy (patch size of 8, 1M updates)  & \textbf{5.60} & \textbf{196.2} & \textbf{408} \\
\midrule

\multirow{1}{*}{$1024\by 1024$}
&RIN+our strategy (patch size of 8, 910K updates)  & \textbf{8.72} & \textbf{163.9} & \textbf{412} \\

\bottomrule
\end{tabular}
\end{table}

\subsection{Visualization of generated samples}

\begin{figure}[h!]
\begin{center}  
\includegraphics[width=1\linewidth]{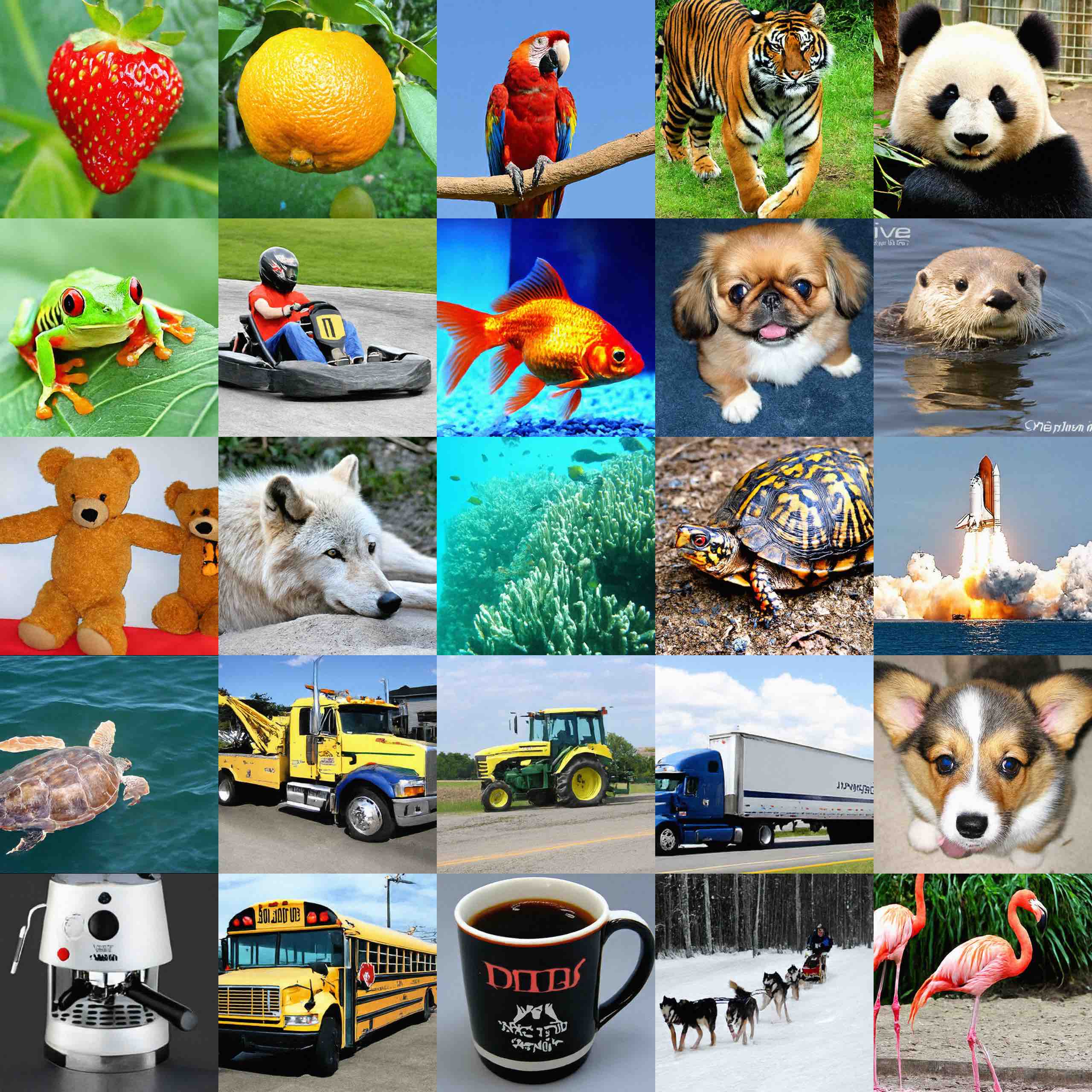}
\end{center}
\caption{\label{fig:samples_512} Random samples at 512$\by$512 resolution generated by our single-stage end-to-end model (trained on class-conditional ImageNet images). The classes are strawberry (949), orange (950), macaw (88), tiger (292), panda (388), tree frog (31), go-kart (573), goldfish (1), pekinese (154), otter (360), teddy bear (850), arctic wolf (270), coral reef (973), box tortoise (37), space shuttle (812), loggerhead sea turtle (33), tow truck (864), tractor (866), trailer truck (867), Pembroke Welsh corgi (263), espresso maker (550), school bus (779), coffee mug (504), dog sled (537), flamingo (130).}
\end{figure}

\begin{figure}[h!]
\begin{center}  
\includegraphics[width=1\linewidth]{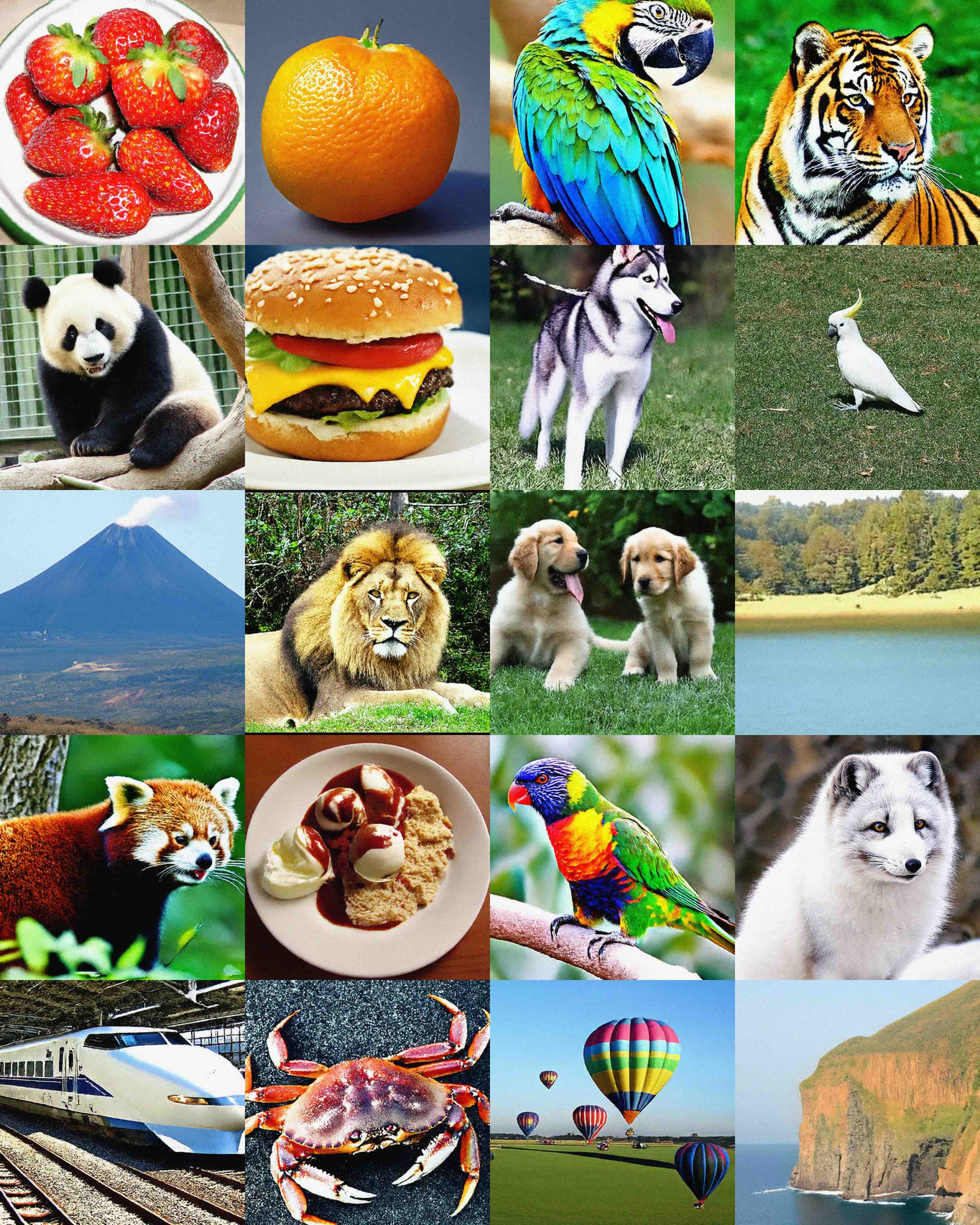}
\end{center}
\caption{\label{fig:samples_768} Random samples at 768$\by$768 resolution generated by our single-stage end-to-end model (trained on class-conditional ImageNet images). The classes are strawberry (949), orange (950), macaw (88), tiger (292), panda (388), cheeseburger (933), husky (250), sulphur-crested cockatoo (89), volcano (980), lion (291), golden retriever (207), lake shore (975), red panda (387), ice cream (928), lorikeet (90), arctic fox (279), bullet train (466), dungeness crab (118), balloon (417), cliff drop-of (972).}
\end{figure}

\begin{figure}[h!]
\begin{center}  
\includegraphics[width=1\linewidth]{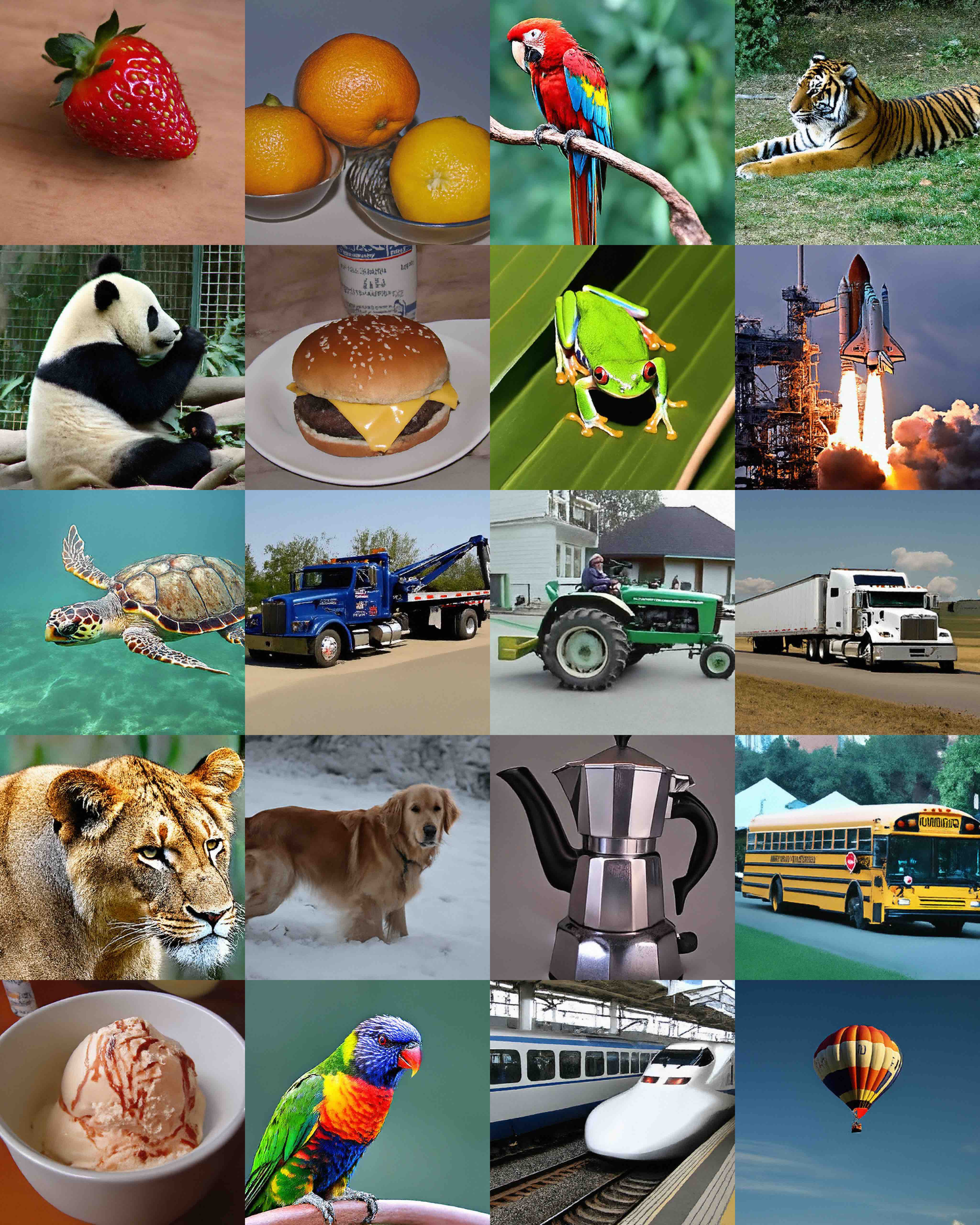}
\end{center}
\caption{\label{fig:samples_1024} Random samples at 1024$\by$1024 resolution generated by our single-stage end-to-end model (trained on class-conditional ImageNet images). The classes are strawberry (949), orange (950), macaw (88), tiger (292), panda (388), cheeseburger (933), tree frog (31), space shuttle (812), loggerhead sea turtle (33), tow truck (864), tractor (866), trailer truck (867), lion (291), golden retriever (207), espresso maker (550), school bus (779), ice cream (928), lorikeet (90), bullet train (466), balloon (417).}
\end{figure}

Even though we do not use label dropout for images at resolution of $512\by512$, we still find the classifier-free guidance~\cite{ho2021classifierfree} during sampling improves the fidelity of generated samples. Therefore, we generate all the visualization samples with a guidance weight of 3.
Figure~\ref{fig:samples_512},~\ref{fig:samples_768} and~\ref{fig:samples_1024} show image samples generated from our trained model. Note that these are random samples, without cherry picking, generated conditioned on the given classes. Overall, we do see the global structure is well preserved across various resolutions, though object parts at smaller scale may be imperfect. We believe it can be improved with scaling the model and/or dataset (e.g., with more detailed text descriptions instead of just the class labels), and also the hyper-parameters tuning (as we do not thoroughly tune them for high resolutions).
  \section{Conclusion}
In this work, we empirically study noise scheduling strategies for diffusion models and show their importance. The noise scheduling not only plays an important role in image generation but also for other tasks such as panoptic segmentation~\cite{chen2022generalist}. A simple strategy of adjusting input scaling factor~\cite{chen2022generalist} works well across different image resolutions. When combined with recently proposed RIN architecture~\cite{jabri2022scalable}, our noise scheduling strategy enables single-stage generation of high resolution images. \textbf{For practitioners, our work suggests that it is important to select a proper noise scheduling scheme when training diffusion models for a new task or a new dataset.}

\clearpage
\section*{Acknowledgements}

We thank David Fleet and Allan Jabri for helpful discussions.

{\small
\bibliography{content/ref}
\bibliographystyle{plainnat}}

\end{document}